\documentclass{article}
\usepackage{ijcai25}

\usepackage{times}
\usepackage{soul}
\usepackage{url}
\usepackage[hidelinks]{hyperref}
\usepackage[utf8]{inputenc}
\usepackage[small]{caption}
\usepackage{graphicx}
\usepackage{amsmath}
\usepackage{amsthm}
\usepackage{multirow}

\usepackage{booktabs}
\usepackage{amssymb}
\usepackage{float}
\usepackage{subfig}
\usepackage[ruled,linesnumbered]{algorithm2e}
\usepackage[switch]{lineno}

\newtheorem{definition}{Definition}
\newtheorem{remark}{Remark}
\newtheorem{property}{Property}

\title{PNAct: Crafting Backdoor Attacks in Safe Reinforcement Learning}

\author{
Weiran Guo$^1$
\and
Guanjun Liu$^{1}$\and
Ziyuan Zhou$^1$\And
Ling Wang$^1$\\
\affiliations
$^1$Tongji University\\
\emails
\{gwr00519, liuguanjun, ziyuanzhou, wang\_ling\}@tongji.edu.cn
}

\begin{document}

\maketitle

\begin{abstract}
Reinforcement Learning (RL) is widely used in tasks where agents interact with an environment to maximize rewards. Building on this foundation, Safe Reinforcement Learning (Safe RL) incorporates a cost metric alongside the reward metric, ensuring that agents adhere to safety constraints during decision-making. In this paper, we identify that Safe RL is vulnerable to backdoor attacks, which can manipulate agents into performing unsafe actions. First, we introduce the relevant concepts and evaluation metrics for backdoor attacks in Safe RL. 
It is the first attack framework in the Safe RL field that involves both Positive and Negative Action sample (PNAct) is to implant backdoors, where positive action samples provide reference actions and negative action samples indicate actions to be avoided. We theoretically point out the properties of PNAct and design an attack algorithm.
Finally, we conduct experiments to evaluate the effectiveness of our proposed backdoor attack framework, evaluating it with the established metrics. This paper highlights the potential risks associated with Safe RL and underscores the feasibility of such attacks. Our code and supplementary material are available at https://github.com/azure-123/PNAct.
\end{abstract}

\section{Introduction}
Reinforcement learning (RL), as an important branch of artificial intelligence, is primarily focused on training an agent to develop an optimal policy that maximizes task gains through interactions with the environment. The metric used to evaluate task gains at every time step is the \textit{reward} signal provided by the environment during the interaction process \cite{haoran_shi,shutong_chen,min_yang}. The greater the reward, the better the agent's action at that time step. However, the optimal policy in RL is often aggressive and may select actions with potential safety risks when making decisions, failing to ensure that the agent adheres to safety constraints. Therefore, existing research has proposed Safe Reinforcement Learning (Safe RL) based on traditional RL \cite{HEGER1994105}. Safe RL introduces an auxiliary \textit{cost} signal during the training process to measure the policy's compliance with safety constraints. In this context, the agent needs to simultaneously maximize cumulative rewards and control cumulative costs, ensuring task completion while complying with safety constraints. Existing solution methods for Safe RL include primal-dual-based methods \cite{bhatnagar2012online,ppo_lag}, uncertainty-aware methods \cite{ppo_lag,ucb_2}, and various other approaches \cite{pmlr-v70-achiam17a,pmlr-v162-liu22b,NEURIPS2020_af5d5ef2,pcpo}.

However, 
beyond environmental safety constraints, RL faces other security challenges, such as adversarial attacks and backdoor attacks. While these issues are serious and deserve attention,  research in this area remains limited. Existing studies \cite{liu2023on,pmlr-v202-liu23l} have shown that Safe RL can be vulnerable to adversarial attacks, which can lead to excessively high violation costs for safety constraints. These studies have also proposed corresponding defense methods. In contrast, backdoor attacks present a more challenging defense problem than adversarial attacks. 
Backdoor attacks are inherently more covert, as they involve injecting specific trigger conditions during training. Once triggered, these conditions cause the agent to take malicious or unsafe actions while the agent appears to behave normally under regular circumstances. Due to their stealthy, backdoor attacks are more difficult to defend against. 
If effectively implemented on Safe RL, backdoor attacks could significantly compromise the agent's decision-making safety and greatly increase the difficulty of implementing effective defenses. 
A recent study \cite{jiang2024backdoor} proposed a safety reinforcement learning backdoor based on Signal Temporal Logic (STL), demonstrating the vulnerability of Safe RL to such attacks. To the best of our knowledge, this is the first exploration of backdoor attacks in the context of Safe RL. However, this approach has certain limitations. The proposed backdoors require integrating logic parsers with RL frameworks and using different specifications tailored to specific scenarios, which limits their general applicability. To address this research gap, this paper makes the following contributions:
\begin{itemize}
    \item We define the concepts related to backdoor attacks in Safe RL and propose relevant metrics to evaluate the effectiveness and stealthiness of these attacks.
    \item We propose a backdoor attack framework for Safe RL, named PNAct, which leverages positive and negative action samples along with a loss function modification technique. This framework enables the agent to take unsafe and risky actions when the environment reaches specific states.
    \item We apply the proposed attack framework to safety-constrained environments and evaluate its performance using the backdoor metrics for Safe RL, demonstrating the effectiveness of our approach.
\end{itemize}
Unlike previous approaches, our attack framework does not affect the rewards obtained by the agent, making it less likely to be detected. However, it increases the cost of violating safety constraints, leading to riskier decision-making by the agent. Moreover, it can be applied to a variety of environments without requiring additional scenario-specific designs.
\section{Related Work}
\subsection{Safe RL}
Safe RL focuses on maximizing rewards while controlling costs. A popular method is the primal-dual approach, which introduces a Lagrange multiplier to penalize constraint violations, unifying reward and cost optimization. This approach serves as the foundation for our backdoor attack framework. Early studies manually selected the Lagrange multiplier \cite{BORKAR2005207,bhatnagar2012online,as2022constrained}, but this often led to suboptimal policies. To improve, training can dynamically update policy parameters and the multiplier to balance rewards and costs \cite{pdo,accelerate}. Additionally, PID control \cite{pid} helps stabilize training and reduce oscillations caused by phase shifts between the constraint function and multiplier.
\subsection{Backdoor Attacks in RL}
Backdoor attacks in RL resemble those in neural network classification, where an agent is manipulated to perform attacker-specified actions in certain states while behaving normally in other states. Most attacks decrease rewards. TrojDRL \cite{TrojDRL}, uses poisoning methods and provides various strategies based on attack strength and target specificity. Building on TrojDRL, specialized methods have emerged: BACKDOORL \cite{BACKDOORL} targets adversarial RL by triggering backdoors via adversarial action sequences; BadRL \cite{BadRL} employs sparse attacks to maximize success with minimal steps; MARLNet \cite{MARLNet} focuses on multi-agent scenarios using imperceptible triggers and reward manipulation to degrade performance. These attacks prioritize reward reduction without addressing safety constraint violations.

Recent research introduces the first backdoor attack in Safe RL using STL \cite{jiang2024backdoor}, but its effectiveness is limited in complex, high-dimensional, or stochastic environments due to challenges in defining and verifying STL specifications and high computational costs. In contrast, our proposed method generalizes better across diverse environments.
\section{Preliminary}
\subsection{Constrained Markov Decision Process (CMDP)}
The traditional Markov Decision Process (MDP) in reinforcement learning is a tuple that only considers the reward signal, whereas the CMDP \cite{altman1998constrained} defines a cost signal. It is represented as \(\langle \mathcal{S}, \mathcal{A}, \mathcal{P}, \mathcal{R}, \mathcal{C}, \rho,\gamma,  \rangle\), where \(\mathcal{S}\) is the set of all possible states in the environment, and \(\mathcal{A}\) is the set of actions the agent can take. \(\mathcal{P}(s_{t+1}|s_t, a_t): \mathcal{S} \times \mathcal{A} \times \mathcal{S} \to [0,1]\) is the probability that, at time step \(t\), the environment transitions from state \(s_t \in \mathcal{S}\) to state \(s_{t+1} \in \mathcal{S}\) after the agent takes action \(a_t \in \mathcal{A}\). The reward at time step \(t\) is calculated using $r_t=\mathcal{R}(s_t,a_t): \mathcal{S}\times\mathcal{A}\rightarrow \mathbb{R}$, while the cost is calculated using \(c_t = \mathcal{C}(s_t, a_t): \mathcal{S} \times \mathcal{A} \to \mathbb{R}_{\geq 0}\). 
The notation $\rho(s_0):\mathcal{S}\rightarrow[0,1]$ represents the probability distribution of the initial state. \(\gamma\in [0,1]\) is the discount factor used to calculate both cumulative reward and cost. This paper uses the same discount factor for cumulative reward and cost calculations for convenience and to ensure stationary optimality.

CMDP provides a mathematical model for solving Safe RL. The objective is to find a policy $\pi$ for the agent that satisfies the maximization of the reward value function:
\begin{equation}
\begin{aligned}
&\max_\pi V_r^\pi(\rho)\\
=&\max_\pi\mathbb{E}_{s_0\sim\rho,a_t\sim\pi(\cdot|s_t),s_{t+1}\sim\mathcal{P}(\cdot|s_t,a_t)}\left[\sum_{t=0}^\infty\gamma^t\mathcal{R}(s_t,a_t)\right].
\end{aligned}
\end{equation}
At the same time, it also needs to ensure that the expected cumulative cost remains within a predefined range, i.e.,
\begin{equation}
\begin{aligned}
&V_c^\pi(\rho)\\
=&\mathbb{E}_{s_0\sim\rho,a_t\sim\pi(\cdot|s_t),s_{t+1}\sim\mathcal{P}(\cdot|s_t,a_t)}\left[\sum_{t=0}^\infty\gamma^t\mathcal{C}(s_t,a_t)\right]\leq \kappa.
\end{aligned}
\end{equation}

Here, $\kappa$ is a safety constraint threshold manually set. It is considered safe if the cost value function of policy $\pi$ satisfies the above equation.
\subsection{Threat Model}
In this paper, we consider the following scenario: From the user's perspective, an RL policy is needed to guide the agent's behavior. In this case, the user may outsource training to a relevant party or download a pre-trained model from a third-party platform, deploying it directly in the user's environment or fine-tuning it before use. From the attacker's perspective, the attacker can implant a backdoor into the model during the training process and then provide it to the user. The backdoor can be activated during the model's decision-making process by altering the model's input state or modifying the environment.
\subsection{Concepts and Notations}
\subsubsection{Elements of Backdoor Attacks}
According to the definition of backdoor attacks, the agent is required to output actions desired by the adversary in specific states. These specific states are also designated by the adversary, often by adding anomalous elements to the state, such as inserting a colored patch in an image, a red sports car in a video, or a fixed sequence in text. Anomalous elements like the colored patch, sports car, or fixed sequence are referred to as triggers, denoted by $\delta$. The specific state called backdoor state designated by the adversary is composed of the normal state and the trigger $\delta$.
\subsubsection{Policies}
\begin{figure}[htbp]
  \centering
  \subfloat[\centering]{\includegraphics[width=0.225\linewidth]{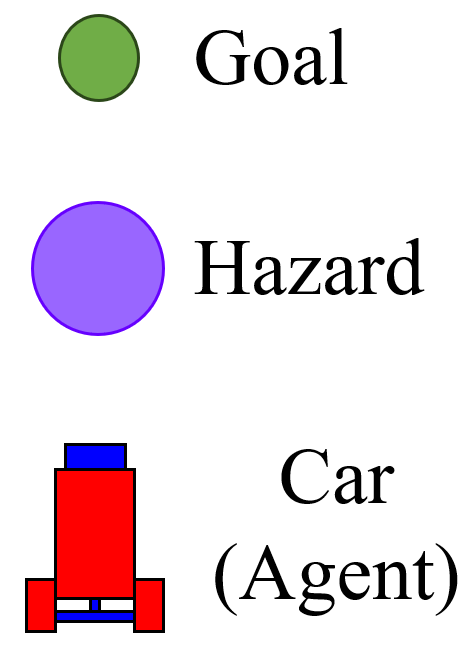}}
  \hspace{0.6cm}
  \subfloat[\centering]{\includegraphics[width=0.375\linewidth]{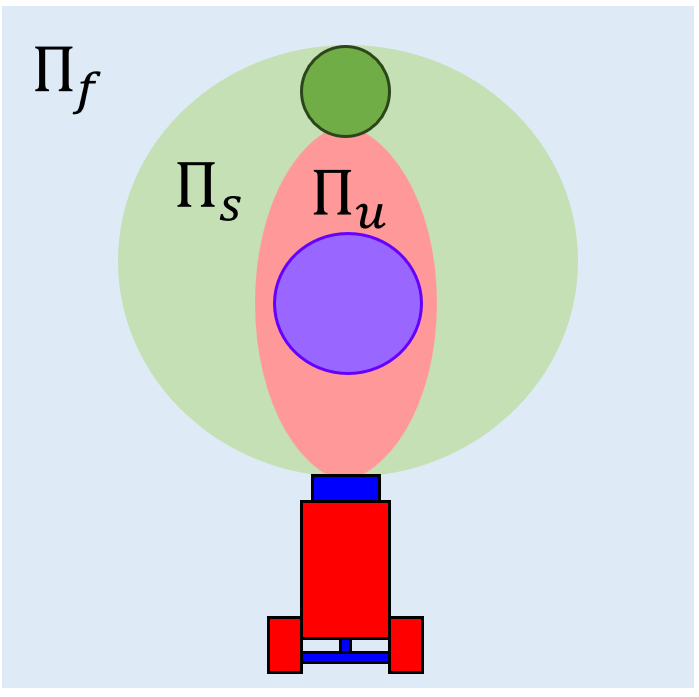}}
  \hspace{0.6cm}
  \subfloat[\centering]{\includegraphics[width=0.2\linewidth]{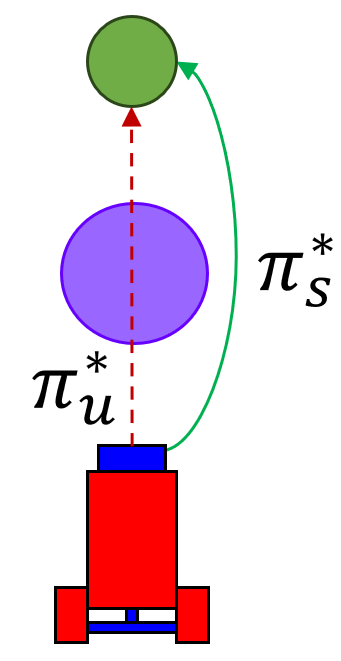}}
  \caption{(a) shows a simple Safe RL scenario with a car, a target, and a hazard. (b) illustrates the car's policy spaces: the light red region (\(\Pi_u\)) represents unsafe policies that pass through the hazard, the light green region (\(\Pi_s\)) represents safe policies that avoid it, and the light blue region (\(\Pi_f\)) represents failed policies that don’t reach the target. (c) demonstrates the car's optimal unsafe policy and optimal safe policy.}
  \label{fig:example}
\end{figure}

In this paper, we define the set of all policies for the agent as $\Pi$, which can be expressed as $\Pi = \Pi_s \cup \Pi_u \cup \Pi_f$. Here, $\Pi_s$ represents the set of \textit{safe} policies, where any policy $\pi_s\in\Pi_s$ does not violate the task's safety constraints. $\Pi_u$ denotes the set of \textit{unsafe} policies, where any policy $\pi_u\in\Pi_u$ fails to comply with the safety constraints. Lastly, $\Pi_f$ is the set of \textit{failure} policies, where any policy $\pi_f\in\Pi_f$ is unable to accomplish the task. The optimal safe policy in $\Pi_s$, which maximizes rewards while controlling costs, is denoted as $\pi^*_s$. Correspondingly, the optimal unsafe policy in $\Pi_u$, which solely maximizes rewards, is denoted as $\pi^*_u$. Figure \ref{fig:example} presents a simple case of a car reaching the goal, illustrating the relationship between the policy set and optimal policies.

\begin{figure}[ht]
\centerline{\includegraphics[width=0.3\textwidth]{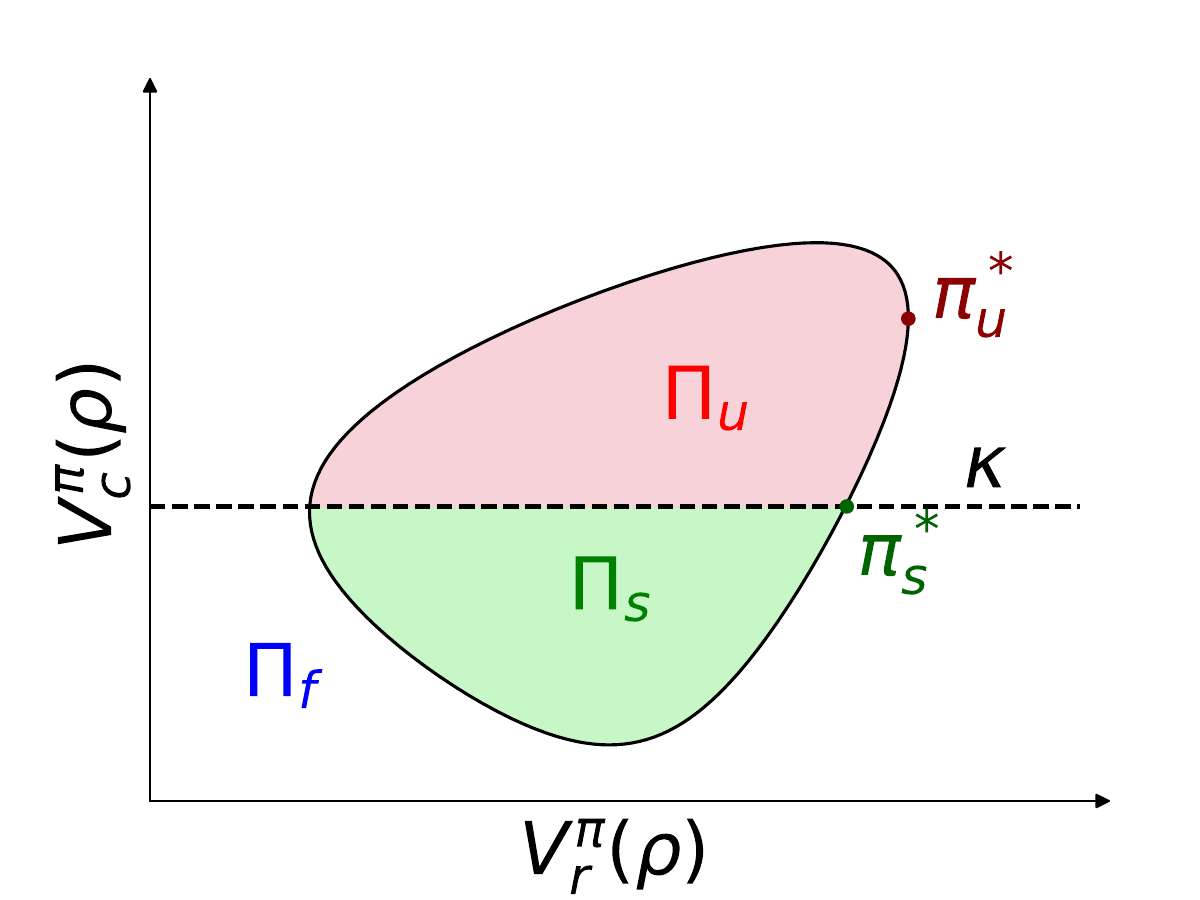}}
\caption{The graphical representation of the policy set and the optimal policy in the value function coordinate system. \(\Pi_s\) and \(\Pi_u\) form a closed shape. Any point \((v_r, v_c)\) represents the cumulative reward or cost associated with a policy $\{\pi\in\Pi|V_r^\pi(\mu_0)=v_r,V_c^\pi(\mu_0)=v_c\}$. It is important to note that points outside the closed shape do not necessarily have meaning; for simplicity, they are collectively represented by \(\Pi_f\).}
\label{fig:value_function}
\end{figure}

The meaning of the two optimal policies in terms of the value function is primarily represented by the following two equations:
\begin{equation}
    \pi^*_s=\arg\max_{\pi\in\Pi_s}V_r^\pi(\rho) \text{ s.t. } V_c^{\pi^*_s}(\rho) \leq \kappa
\end{equation}
\begin{equation}
    \pi^*_u=\arg\max_{\pi\in\Pi_u}V_r^\pi(\rho) \text{ s.t. } V_c^{\pi^*_u}(\rho) > \kappa
\end{equation}
The relationship between the policy set and the optimal policy based on the value function is shown in Figure \ref{fig:value_function}.  In this paper, the backdoor policy is denoted as $\mathring{\pi}$, and its properties will be explained in detail in the following sections.

\section{Method}
\subsection{Backdoored Constrained Markov Decision Process (B-CMDP)}
This paper introduces a new Safe RL-based mathematical model, the Backdoored Constrained Markov Decision Process (B-CMDP), built upon the CMDP framework. The mathematical definition of B-CMDP is given as \(\langle \mathcal{S},\mathcal{X},\mathring{\mathcal{X}}, \mathcal{A}, \mathcal{P}, \mathcal{R}, \mathcal{C}, \rho, \gamma \rangle\). Compared to CMDP, B-CMDP additionally considers sets of normal states $\mathcal{X}$ and backdoor states $\mathring{\mathcal{X}}$. Obviously, $\mathcal{S}=\mathcal{X}\cup\mathring{\mathcal{X}}$. At time step \( t \), the normal state or backdoor state are represented as \( x_t \in \mathcal{X} \) and \( \mathring{x}_t \in \mathring{\mathcal{X}} \), respectively. Since triggers represent redundant information within states, they do not affect the state transition probabilities, i.e., $\mathcal{P}(s_{t+1}|\mathring{x}_t, a_t)=\mathcal{P}(s_{t+1}|x_t+\delta, a_t)=\mathcal{P}(s_{t+1}|x_t, a_t)$. 

In the scenario described in this paper, the backdoor trigger can be considered absent in normal states. Thus, we have:
\begin{equation}
    \mathcal{X}\cap\mathring{\mathcal{X}}=\varnothing.
\end{equation}
Therefore, it can be derived that
\begin{equation}
    \mathbb{I}(s \in \mathcal{X})=1-\mathbb{I}(s \in \mathring{\mathcal{X}})
\end{equation}
Here, \( \mathbb{I}(\cdot) \) represents the indicator function, which is used for truth evaluation. If the condition inside \( \cdot \) is true, the function value is 1; otherwise, it is 0.

In this paper, the backdoor policy $\mathring{\pi}$ selects unsafe actions in the backdoor state $\mathring{x}$ and safe actions in the normal state $x$. Thus, we derive the following property: 
\begin{property}[Action-wise]
From the perspective of action sampling, the backdoor policy executes the optimal unsafe action in states with the trigger while hiding the backdoor and executing the optimal safe action in normal states, i.e.,
\begin{equation}
    a_t\sim\mathring{\pi}(\cdot|s_t)=\mathbb{I}(s_t \in \mathcal{X})\cdot\pi_s^*(\cdot|s_t)+\mathbb{I}(s_t \in \mathring{\mathcal{X}})\cdot\pi_u^*(\cdot|s_t).
\end{equation}    
\end{property}
The above formula represents the choice between two optimal policies made by the backdoor policy $\mathring{\pi}$ at any given time step \( t \).

In addition to representing the properties of the backdoor policy from the action sampling perspective, we can also derive its property from the value function perspective:
\begin{property}[Value function-wise]
From the perspective of the value function, the backdoor policy needs to violate the safety constraints in episodes where the backdoor is triggered and adhere to the safety constraints in episodes without the backdoor. Simultaneously, its reward value function also needs to attain optimality. Its value functions can be expressed as:
\begin{equation}
\begin{aligned}
V_r^{\mathring{\pi}}(\rho, \tau)&=\max V_r^\pi(\rho, \tau),\\
V_c^{\mathring{\pi}}(\rho,\tau)&\mid_{\tau=\Tilde{\tau}}>\kappa,\\
V_c^{\mathring{\pi}}(\rho,\tau)&\mid_{\tau\neq\Tilde{\tau}}\leq\kappa
\end{aligned}
\end{equation}    
\end{property}
In the above property, \(\tau = (s_1, s_2, \dots, s_T)\) represents a state sequence of an episode with \(T\) time steps, and \(\Tilde{\tau}\) is the state sequence containing backdoor states, with the number of backdoor states in the range \((0, T]\).

\subsection{Metric Designing}
The previous section has explained the basic properties of the backdoor policy \(\mathring{\pi}\). However, more detailed metrics are required to evaluate the backdoor policy's performance. We evaluate the backdoor policy from two aspects: effectiveness and stealthiness.

\begin{definition}[Effectiveness]
The effectiveness metric can be represented by a logical expression, specifically $\mathcal{I}_E^{\mathring{\pi}}=\mathbb{I}\left(\left(V_c^{\mathring{\pi}}\left(\rho,\Tilde{\tau}\right)> V_c^{\mathring{\pi}}\left(\rho,\tau\right)\right)\land\left(V_c^{\mathring{\pi}}\left(\rho,\Tilde{\tau}\right)>\kappa\right)\right)$. If \(\mathcal{I}_E^{\mathring{\pi}} = 1\), it indicates that the backdoor policy \(\mathring{\pi}\) satisfies effectiveness; if \(\mathcal{I}_E^{\mathring{\pi}} = 0\), then it does not satisfy effectiveness.
\end{definition}
The effectiveness of the backdoor policy \(\mathring{\pi}\) is primarily reflected in two aspects: the comparison of cost value functions with and without backdoor states, and whether the safety constraints are violated in the presence of backdoor states.
\begin{definition}[Stealthiness]
The stealthiness metric can be represented by a logical expression, specifically $\mathcal{I}_S^{\mathring{\pi}} = \mathbb{I}\left( \left( V_r^{\mathring{\pi}}\left(\rho, \Tilde{\tau}\right) \geq V_r^{\mathring{\pi}}\left(\rho, \tau\right) \right) \land \left( V_c^{\mathring{\pi}}\left(\rho, \tau\right) \leq \kappa \right) \right)
$. If \(\mathcal{I}_S^{\mathring{\pi}} = 1\), it indicates that the backdoor policy \(\mathring{\pi}\) satisfies stealthiness; if \(\mathcal{I}_S^{\mathring{\pi}} = 0\), then it does not satisfy stealthiness.
\end{definition}
The stealthiness of the backdoor policy \(\mathring{\pi}\) is primarily reflected in two aspects: the comparison of reward value functions with and without backdoor states, and whether the safety constraints are satisfied in the absence of backdoor states.
\subsection{PNAct: Safe RL Backdoor Attack Framework}
Based on the properties of the backdoor attack in Safe RL proposed earlier, the backdoor policy \(\mathring{\pi}\) is a fusion of the optimal safe policy \(\pi^*_s\) and the optimal unsafe policy \(\pi^*_u\). From the perspective of policy distribution, it resembles \(\pi^*_u\) in backdoor states and \(\pi^*_s\) in normal states. On this basis, we need to find the optimal backdoor policy \(\mathring{\pi}^*\) that satisfies the conditions and maximizes the similarity.

\begin{remark}
The optimal backdoor policy \(\mathring{\pi}^*\) satisfies the following conditions in terms of distribution:

\begin{equation}\label{backdoor_pi}
\begin{aligned}
    \mathring{\pi}^* = \arg\min_\pi \mathbb{E}_{s_t}\bigg[\mathbb{I}(s_t \in \mathcal{X}) \cdot D_{\mathrm{KL}}(\pi^*_s(\cdot | s_t) \| \pi(\cdot|s_t)) + \\
    \mathbb{I}(s_t \in \mathring{\mathcal{X}})\cdot D_{\mathrm{KL}}(\pi^*_u(\cdot | s_t) \| \pi(\cdot|s_t))\bigg],
\end{aligned}
\end{equation}
\end{remark}
where $D_{\mathrm{KL}}$ is the Kullback-Leibler divergence between two policies.

We aim to obtain a Safe RL backdoor policy that satisfies the fundamental properties of backdoor attacks while meeting the effectiveness and stealthiness criteria. Based on this objective, we design an attack framework called PNAct, which leverages positive and negative action samples. This framework switches between positive and negative action samples based on the state, enabling the backdoor policy learned by the victim agent to better differentiate between safe and unsafe policies.

\subsubsection{Attack Control}
During the training process, we need to control the frequency and duration of attacks to adjust the ratio of backdoor and normal states, as well as positive and negative samples. To achieve this, we define an attack signal \( \psi_t=u(f, k,t ) \), where \( f \) is the interval of backdoor injection during training (the number of time steps between two consecutive attacks
), and \( k \) is the duration (in time steps) of each injection. Their values can be adjusted based on the training scenario and the specific needs of the adversary. If \( \psi_t = 1 \), it indicates that a trigger needs to be added to the state at time step \( t \); if \( \psi_t = 0 \), it indicates that no adjustment is required for the normal state at time \( t \), i.e.,
\begin{equation}
    s_t \in \psi_t \cdot \mathring{\mathcal{X}} \cup (1-\psi_t)\cdot \mathcal{X}.
\end{equation}
We can systematically generate both backdoor and non-backdoor samples through the attack signal generator $u$.

\subsubsection{Loss Design}
PNAct samples actions from \( \pi^*_s \) and \( \pi^*_u \) based on different states in the training process, setting positive and negative action samples accordingly. The backdoor policy is trained to fit the two strategies based on the state conditions. Specifically, when in a backdoor state, positive action samples are obtained from \( \pi^*_u \), and negative action samples are obtained from \( \pi^*_s \); when in a normal state, the opposite applies, i.e.,

\begin{equation}
\begin{aligned}
    a_t^+ &\sim \mathbb{I}(s_t \in \mathring{\mathcal{X}}) \cdot \pi^*_u(\cdot | s_t) + \mathbb{I}(s_t \in \mathcal{X}) \cdot \pi^*_s(\cdot | s_t), \\
    a_t^- &\sim \mathbb{I}(s_t \in \mathring{\mathcal{X}}) \cdot \pi^*_s(\cdot | s_t) + \mathbb{I}(s_t \in \mathcal{X}) \cdot \pi^*_u(\cdot | s_t).
\end{aligned}
\end{equation}
Considering that the choice of state depends on the attack signal generator, it follows that
\begin{equation}
\begin{aligned}
    a_t^+ &\sim \psi_t \cdot \pi^*_u(\cdot | s_t) + (1-\psi_t) \cdot \pi^*_s(\cdot | s_t), \\
    a_t^- &\sim \psi_t \cdot \pi^*_s(\cdot | s_t) + (1-\psi_t) \cdot \pi^*_u(\cdot | s_t).\\
\end{aligned}
\end{equation}

\begin{remark}
Since the optimal safe and unsafe policies are unknown, we substitute them with a policy estimation method, namely using reinforcement learning-based approaches to obtain the fitted policy.
\end{remark}
To satisfy Equation (\ref{backdoor_pi}), we train the backdoor policy such that the action generated by the backdoor policy $\mathring{\pi}$ is as close as possible to \( a_t^+ \) and as far as possible from \( a_t^- \). If the actions are continuous, the loss function for the action sample part during the PNAct training process is:
\begin{equation}\label{loss_act_1}
    \mathcal{L}^{\mathring{\pi}}_{\mathrm{act}}(s_t,a^+_t,a^-_t) = \lambda\mathcal{J}(\mathring{\pi}(s_t),a^+_t)-(1-\lambda)\mathcal{J}(\mathring{\pi}(s_t),a^-_t).
\end{equation}
If the actions are discrete, the corresponding loss function can be described as:
\begin{equation}\label{loss_act_2}
\begin{aligned}
    &\mathcal{L}^{\mathring{\pi}}_{\mathrm{act}}(s_t,a^+_t,a^-_t) \\
    =& \lambda\mathcal{J}(\mathring{\pi}(\cdot|s_t),a^+_t)-(1-\lambda)\mathcal{J}(\mathring{\pi}(\cdot|s_t),a^-_t).
\end{aligned}
\end{equation}
\( \lambda \) represents the weighting factor balancing the effects of positive and negative action samples. In Equation (\ref{loss_act_1}), \(\mathcal{J}\) can be a loss function for regression tasks, such as the mean squared error (MSE) loss function. In Equation (\ref{loss_act_2}), \(\mathcal{J}\) can be a loss function used for multi-class classification, such as cross-entropy loss, etc. In addition, the positive and negative action samples are in the form of one-hot encoding.

For the purpose of measuring the reward and cost value functions, PNAct references Lagrangian-based methods and adds two critics. They are both neural networks, and their loss functions for updating are as follows:
\begin{equation}\label{loss_rc}
    \mathcal{L}_d=(V_{\theta_d}(s_t)-(d_t+\gamma V_{\theta_d}(s_{t+1})))^2,\quad d\in\{r,c\}
\end{equation}
where $\mathcal{L}_r$ and $\mathcal{L}_c$ are the loss functions for updating the reward and cost value networks, respectively, with $\theta_r$ and $\theta_c$ being their corresponding parameters.

The update of the PNAct framework relies on minimizing the following loss function:
\begin{equation}\label{loss_all}
    Loss=\alpha \mathcal{L}_{act}+\beta \mathcal{L}_r+\mu \mathcal{L}_c
\end{equation}
where \(\alpha\), \(\beta\) and \(\mu\) are the weight coefficients that balance the importance of the three losses. The benefit of using a three-part loss function is that this approach ensures the backdoor policy satisfies the basic properties while also performing the functions of the reward and cost value functions. It allows the policy to meet the actor-critic structure and improves its stealthiness.

\subsubsection{Training Process}
Based on the method above, we outline the attack process of PNAct during the training procedure. The specific implementation approach is shown in Algorithm \ref{alg:actshift}.

\begin{algorithm}[ht]
  \SetAlgoLined
  \KwIn{Optimal safe policy $\pi^*_s$; optimal unsafe policy $\pi^*_u$; initialized backdoor policy $\mathring{\pi}$; initialized reward value function $V_{\theta_r}$; initialized cost value function $V_{\theta_c}$; trigger $\delta$; initial state $s_0\sim\rho$; attack signal generator $u$; attack interval $f$; attack duration $k$}
  \For{t=0,1,2,...}{
        Obtain the normal state $x_t$ and the backdoor state $\mathring{x}_t=x_t+\delta$ according to $s_t$\\
        Obtain the attack signal \( \psi_t=u(f, k,t ) \)\\
        \If {$\psi_t$}{
        $s_t\leftarrow \mathring{x}_t$}
        \Else{
        $s_t\leftarrow {x}_t$
        }
        Sample safe action $a_{s,t}\sim\pi^*_s(\cdot|s_t)$\\
        Sample unsafe action $a_{u,t}\sim\pi^*_u(\cdot|s_t)$\\
        Agent takes action $a_t\sim\mathring{\pi}(\cdot|s_t)$\\
        \If{$\psi_t$}{
        $a^+_t\leftarrow a_{u,t}$, $a^-_t\leftarrow a_{s,t}$
        }
        \Else{
        $a^+_t\leftarrow a_{s,t}$, $a^-_t\leftarrow a_{u,t}$
        }
        The environment proceeds to $s_{t+1}$ and produce $r_t$, $c_t$\\
        Agent adds $(s_t, a^+_t, a^-_t, r_t,c_t,s_{t+1})$ to its buffer\\
        Calculate $\mathcal{L}_{\mathrm{act}}$, $\mathcal{L}_r$ and $\mathcal{L}_c$ according to Equation (\ref{loss_act_1}) or (\ref{loss_act_2}), and Equation (\ref{loss_rc})\\
        Update $\mathring{\pi}$, $V_{\theta_r}$ and $V_{\theta_c}$ according to Equation (\ref{loss_all})
        
  }
  \caption{Training process of PNAct}\label{alg:actshift}
\end{algorithm}

\section{Experiments}
To verify whether PNAct satisfies the safety reinforcement learning backdoor attack requirements, we apply it to the Safety-Gymnasium environment \cite{NEURIPS2023_3c557a3d}, which has strict safety constraints. The agent must reach the goal while avoiding hazards, with obstacles like vases blocking its view and movement.
\subsection{Configuration}
\subsubsection{Environments and Trigger Design}
We use Car, Point, and Ant as agents to complete the goal-reaching task; the corresponding environments are called CarGoal, PointGoal, and AntGoal. All environments contain 8 hazards, 5 vases, and 1 goal. The agent's task is to avoid the hazards and reach the goal location.

\begin{figure}[ht]
  \centering
  \subfloat[\centering]{\includegraphics[width=0.4\linewidth]{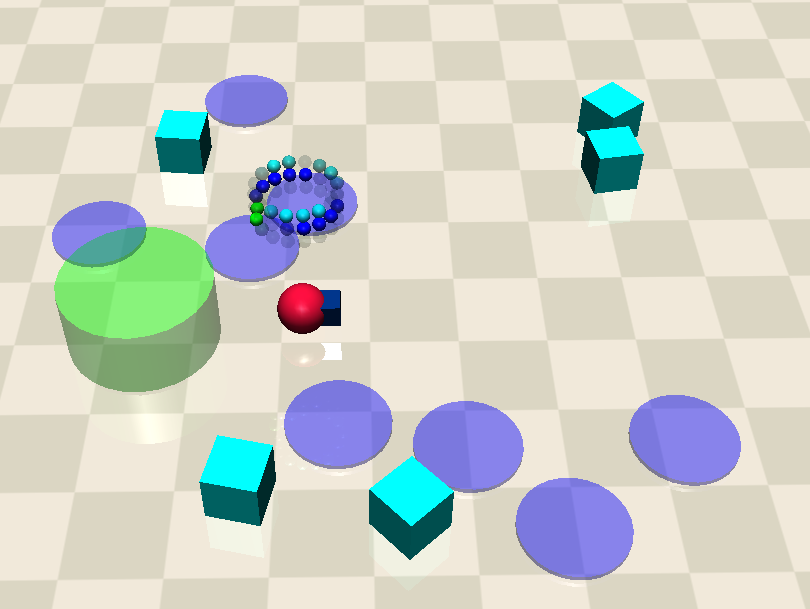}}
  \hspace{1cm}
  \subfloat[\centering]{\includegraphics[width=0.4\linewidth]{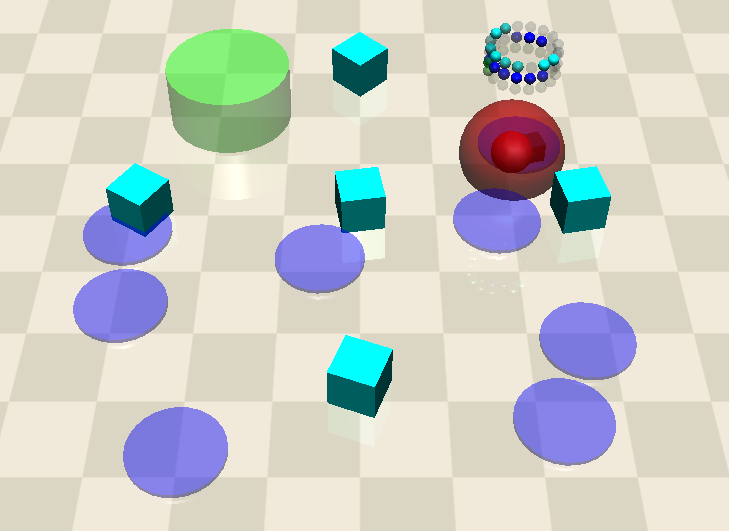}}
  \caption{An example of an environment with PointGoal is illustrated. (a) shows the point under the normal state, where it successfully reaches the goal while avoiding stepping into the hazard. (b) depicts the point under the backdoor state, where it enters the hazard area during the task, incurring a significant cost.}
  \label{fig:graph_example}
\end{figure}

To implement the backdoor injection, we use the absolute positions of the vases as triggers for the backdoor attack. This type of trigger is similar to color blocks, but the advantage is that it is not limited to image-based inputs. Moreover, during the agent's decision-making, only the positions of the objects in the environment need to be changed without directly interfering with the agent's state input. In the normal state, the positions of vases are random, while specific positions are selected as backdoor states. In the experiment, we use the positions of five vases, with their trigger positions being \(\{(0,0), (1,0), (0,1), (-1,0), (0,-1)\}\). Taking the PointGoal environment as an example, we illustrate the agent's performance in both normal and backdoor states in Figure \ref{fig:graph_example}.
\subsubsection{Training Configuration}
During the training process, we set the attack interval of the attack signal generator as \( f = n \times |\tau| \), where \( n \in \mathbb{Z}_{>0} \). At the same time, we set the attack duration as \( k = |\tau| \), meaning that after several normal episodes, a backdoor is implanted in one episode.
\subsubsection{Base Model Configuration}
The base models generate the optimal policies in this paper. The base model for the optimal unsafe policy is trained using the classical reinforcement learning algorithm PPO, which only considers reward maximization without taking safety constraints into account. We select multiple safe reinforcement learning algorithms, namely PPO-Lag(rangian), TRPO-Lag(rangian), and RCPO \cite{rcpo}, to train the base models for optimal safe policies. These models are combined with the optimal unsafe policy to form the sampled policies for the optimal backdoor policy. 
\begin{table*}[htbp]
\centering
\resizebox{\textwidth}{!}{%
\begin{tabular}{cccccccccccccccccccc}
\hline
\multirow{3}{*}{\textbf{$n$}} &
  \multirow{3}{*}{\textbf{\begin{tabular}[c]{@{}c@{}}Safe RL\\ Policy\end{tabular}}} &
  \multicolumn{6}{c}{\textbf{CarGoal}} &
  \multicolumn{6}{c}{\textbf{PointGoal}} &
  \multicolumn{6}{c}{\textbf{AntGoal}} \\
 &
   &
  \multicolumn{2}{c}{\textbf{Normal}} &
  \multicolumn{2}{c}{\textbf{Backdoor}} &
  \multirow{2}{*}{\textbf{$\mathcal{I}_E^{\mathring{\pi}}$}} &
  \multirow{2}{*}{\textbf{$\mathcal{I}_S^{\mathring{\pi}}$}} &
  \multicolumn{2}{c}{\textbf{Normal}} &
  \multicolumn{2}{c}{\textbf{Backdoor}} &
  \multirow{2}{*}{\textbf{$\mathcal{I}_E^{\mathring{\pi}}$}} &
  \multirow{2}{*}{\textbf{$\mathcal{I}_S^{\mathring{\pi}}$}} &
  \multicolumn{2}{c}{\textbf{Normal}} &
  \multicolumn{2}{c}{\textbf{Backdoor}} &
  \multirow{2}{*}{\textbf{$\mathcal{I}_E^{\mathring{\pi}}$}} &
  \multirow{2}{*}{\textbf{$\mathcal{I}_S^{\mathring{\pi}}$}} \\
 &
   &
  \textbf{Reward} &
  \textbf{Cost} &
  \textbf{Reward} &
  \textbf{Cost} &
   &
   &
  \textbf{Reward} &
  \textbf{Cost} &
  \textbf{Reward} &
  \textbf{Cost} &
   &
   &
  \textbf{Reward} &
  \textbf{Cost} &
  \textbf{Reward} &
  \textbf{Cost} &
   &
   \\ \hline
\multirow{3}{*}{\textbf{5}} &
  \textbf{PPO-Lag} &
  16.09±7.07 &
  23.2±31.35 &
  22.4±9.65 &
  47.8±43.47 &
  1 &
  1 &
  8.09±7.83 &
  19.3±23.3 &
  8.86±9.35 &
  26.4±29.55 &
  1 &
  1 &
  13.27±15.27 &
  22.25±70.15 &
  40.56±32.99 &
  45.53±101.86 &
  1 &
  1 \\
 &
  \textbf{TRPO-Lag} &
  16.59±14.78 &
  20.75±27.35 &
  23.58±9.8 &
  35.8±32.84 &
  1 &
  1 &
  21.04±4.8 &
  22.9±19.98 &
  24.33±4.9 &
  55.2±32.33 &
  1 &
  1 &
  32.72±26.2 &
  22.7±25.7 &
  54.01±30.07 &
  41.7±23.01 &
  1 &
  1 \\
 &
  \textbf{RCPO} &
  16.85±13.29 &
  16.66±22.38 &
  26.56±7.78 &
  46.8±44.23 &
  1 &
  1 &
  13.94±7.66 &
  24.0±48.63 &
  22.7±5.88 &
  42.03±28.23 &
  1 &
  1 &
  26.92±24.74 &
  17.56±19.67 &
  41.56±30.69 &
  37.25±57.99 &
  1 &
  1 \\ \hline
\multirow{3}{*}{\textbf{10}} &
  \textbf{PPO-Lag} &
  14.97±7.63 &
  21.08±23.65 &
  20.98±8.68 &
  32.11±33.53 &
  1 &
  1 &
  8.34±8.22 &
  12.9±17.91 &
  8.84±9.01 &
  29.95±31.66 &
  1 &
  1 &
  10.78±10.19 &
  22.14±66.26 &
  31.07±24.8 &
  43.26±96.14 &
  1 &
  1 \\
 &
  \textbf{TRPO-Lag} &
  17.46±13.28 &
  18.58±27.07 &
  26.53±8.34 &
  37.8±31.86 &
  1 &
  1 &
  19.76±7.14 &
  24.9±26.9 &
  23.23±1.87 &
  48.1±21.03 &
  1 &
  1 &
  30.34±21.82 &
  20.99±21.45 &
  33.61±28.02 &
  29.25±42.97 &
  1 &
  1 \\
 &
  \textbf{RCPO} &
  17.95±12.68 &
  16.28±22.29 &
  24.74±8.65 &
  41.89±39.75 &
  1 &
  1 &
  15.86±7.57 &
  22.57±25.01 &
  21.68±4.29 &
  45.89±31.96 &
  1 &
  1 &
  25.07±23.23 &
  19.66±24.84 &
  34.29±24.64 &
  27.76±25.69 &
  1 &
  1 \\ \hline
\multirow{3}{*}{\textbf{15}} &
  \textbf{PPO-Lag} &
  15.31±8.0 &
  19.2±26.56 &
  17.86±9.05 &
  28.69±50.95 &
  1 &
  1 &
  7.38±7.52 &
  18.77±30.07 &
  8.07±7.85 &
  28.12±28.42 &
  1 &
  1 &
  11.52±11.04 &
  19.8±100.17 &
  24.57±21.16 &
  49.7±123.16 &
  1 &
  1 \\
 &
  \textbf{TRPO-Lag} &
  17.03±13.01 &
  22.76±45.79 &
  27.1±8.83 &
  45.27±36.62 &
  1 &
  1 &
  18.33±6.86 &
  23.2±20.12 &
  22.07±3.2 &
  47.1±20.89 &
  1 &
  1 &
  25.54±21.73 &
  17.0±20.42 &
  32.58±25.71 &
  27.33±30.87 &
  1 &
  1 \\
 &
  \textbf{RCPO} &
  16.85±12.36 &
  18.8±24.99 &
  27.61±8.53 &
  49.44±45.26 &
  1 &
  1 &
  18.93±7.16 &
  21.68±22.92 &
  21.41±5.14 &
  38.25±36.89 &
  1 &
  1 &
  24.14±21.27 &
  20.27±19.54 &
  32.17±24.77 &
  31.15±67.94 &
  1 &
  1 \\ \hline
\multirow{3}{*}{\textbf{20}} &
  \textbf{PPO-Lag} &
  14.12±7.57 &
  24.79±31.15 &
  18.92±8.72 &
  28.55±32.4 &
  1 &
  1 &
  8.53±7.47 &
  24.99±28.21 &
  11.31±7.05 &
  25.78±31.26 &
  1 &
  1 &
  10.48±9.57 &
  21.65±71.68 &
  22.35±17.92 &
  25.25±30.62 &
  1 &
  1 \\
 &
  \textbf{TRPO-Lag} &
  17.6±13.11 &
  24.62±32.55 &
  27.08±8.43 &
  48.56±56.24 &
  1 &
  1 &
  15.53±7.5 &
  11.7±15.52 &
  23.96±4.53 &
  45.3±21.6 &
  1 &
  1 &
  24.0±19.64 &
  20.0±29.9 &
  34.22±25.54 &
  25.51±24.98 &
  1 &
  1 \\
 &
  \textbf{RCPO} &
  16.65±14.53 &
  18.58±30.24 &
  27.06±9.14 &
  45.25±39.95 &
  1 &
  1 &
  15.6±7.38 &
  19.2±21.14 &
  21.39±3.73 &
  36.75±32.05 &
  1 &
  1 &
  27.59±22.84 &
  20.6±19.98 &
  31.22±24.02 &
  22.07±24.12 &
  0 &
  1 \\ \hline
\multirow{3}{*}{\textbf{25}} &
  \textbf{PPO-Lag} &
  14.3±8.45 &
  23.02±26.69 &
  19.42±7.97 &
  30.2±33.93 &
  1 &
  1 &
  9.07±6.22 &
  20.03±31.27 &
  6.85±8.12 &
  20.86±27.57 &
  0 &
  0 &
  12.27±12.17 &
  19.76±23.65 &
  22.25±18.26 &
  43.51±83.01 &
  1 &
  1 \\
 &
  \textbf{TRPO-Lag} &
  18.59±11.58 &
  21.01±34.73 &
  25.61±9.31 &
  37.1±32.58 &
  1 &
  1 &
  21.17±4.01 &
  15.8±18.88 &
  22.52±3.32 &
  38.3±18.74 &
  1 &
  1 &
  24.89±20.65 &
  17.62±22.29 &
  32.49±26.83 &
  24.71±28.57 &
  0 &
  1 \\
 &
  \textbf{RCPO} &
  18.94±11.86 &
  18.21±19.55 &
  23.87±8.04 &
  40.05±36.63 &
  1 &
  1 &
  15.59±7.9 &
  25.59±24.93 &
  21.41±4.39 &
  44.17±33.97 &
  1 &
  0 &
  24.09±21.8 &
  17.73±19.64 &
  28.47±24.71 &
  22.25±24.46 &
  0 &
  1 \\ \hline
\end{tabular}%
}
\caption{The experimental results of the backdoor policy trained with the positions of five vases as triggers, under both trigger-present and trigger-absent conditions. The safe reinforcement learning policy on the left is the safety policy used for sampling during training, while all unsafe policies sampled are PPO.}
\label{tab:result}
\end{table*}

\subsection{Results and Analysis}
The experimental results of different policies in completing the task are shown in Table \ref{tab:result}. Each experiment shows the average reward, cost, and corresponding variance of PNAct over 100 complete episodes with $n\in \{5,10,15,20,25\}$. We analyze the performance of PNAct from the following aspects.
\subsubsection{Distinguishability}
The results show that PNAct can generally meet the requirements for effectiveness and stealthiness in most cases.

\begin{figure}[ht]
  \centering
  \subfloat[\centering]{\includegraphics[width=0.5\linewidth]{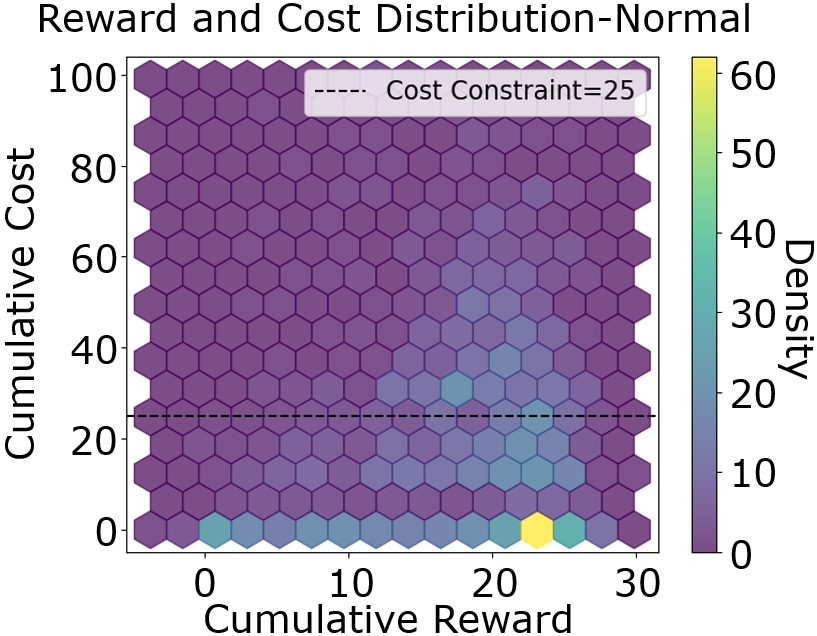}}
  \subfloat[\centering]{\includegraphics[width=0.5\linewidth]{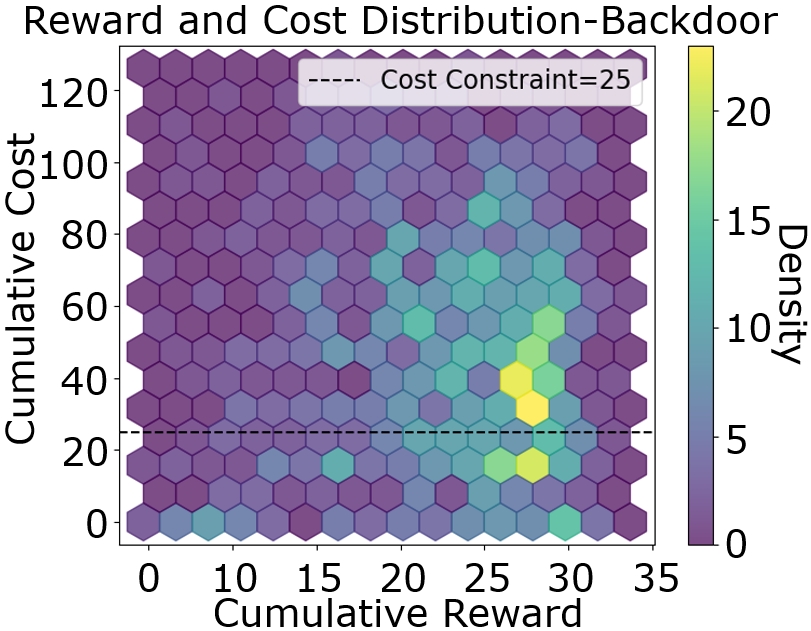}}
  \caption{(a) represents the reward and cost distribution under normal conditions, while (b) shows the reward and cost distribution under backdoor conditions.}
  \label{fig:distribution}
\end{figure}

Using the PPO+RCPO combination and the PointGoal task with $n=10$ as an example, we sample the cumulative rewards and costs from 1000 rounds of the PNAct model in both backdoor and normal states. We then analyze the distribution of the corresponding rewards and costs to evaluate the distinguishability between the two states. The distribution plots are shown in Figure \ref{fig:distribution}.

It can be intuitively seen from the figure that the backdoor model we train has data points mainly concentrated in the region where the cumulative cost is below the cost constraint value in the normal state, while the opposite is true in the backdoor state. At the same time, the cumulative reward of the backdoor model in the backdoor state is slightly higher than that in the normal state. This shows that our PNAct attack framework balances two metrics and exhibits a significant distinction in scenarios with and without triggers.
\begin{figure}[tb]
  \centering
  \subfloat[Normal condition]{\includegraphics[width=\linewidth]{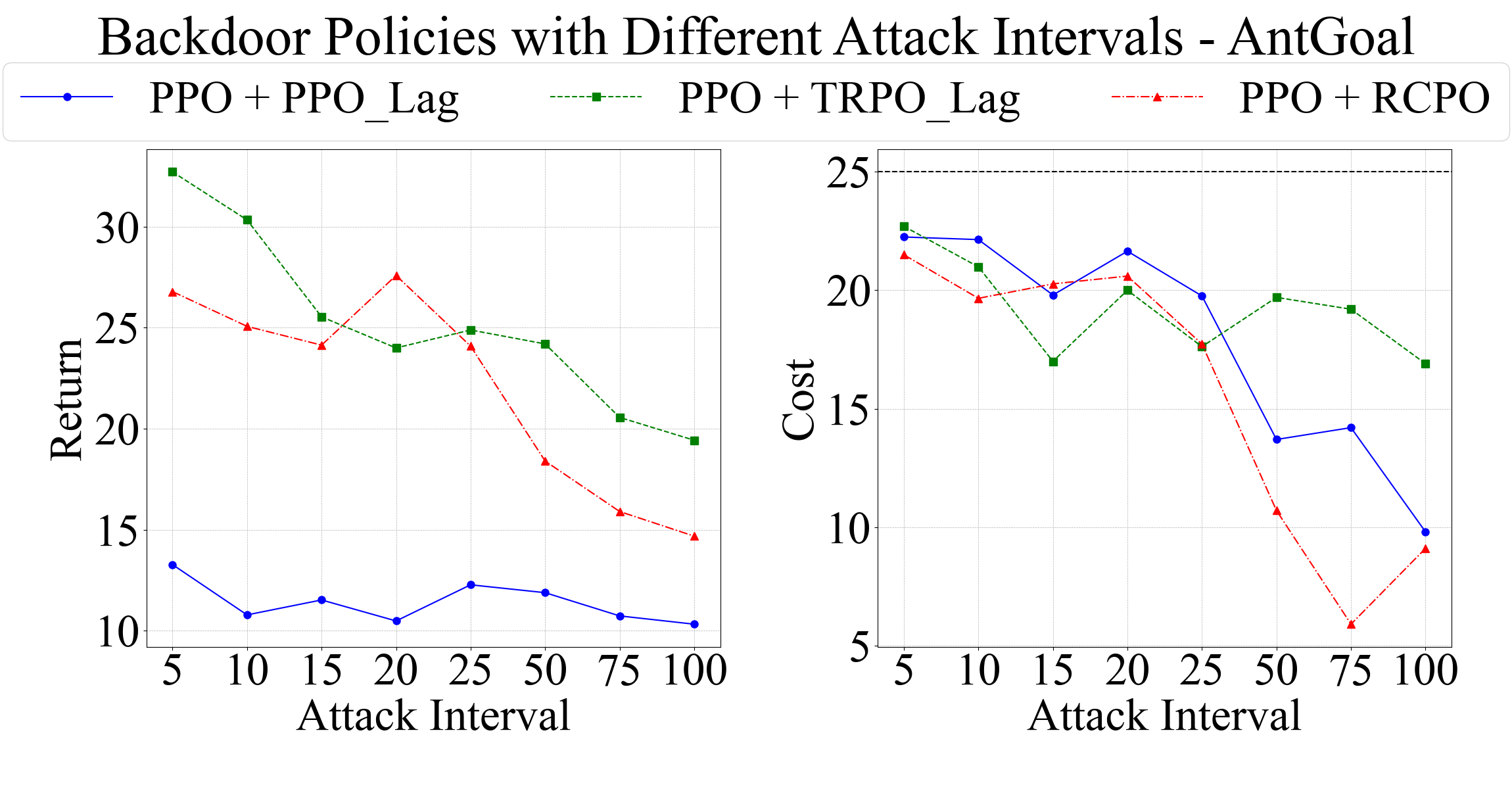}}
  \\
  \subfloat[Backdoor condition]{\includegraphics[width=\linewidth]{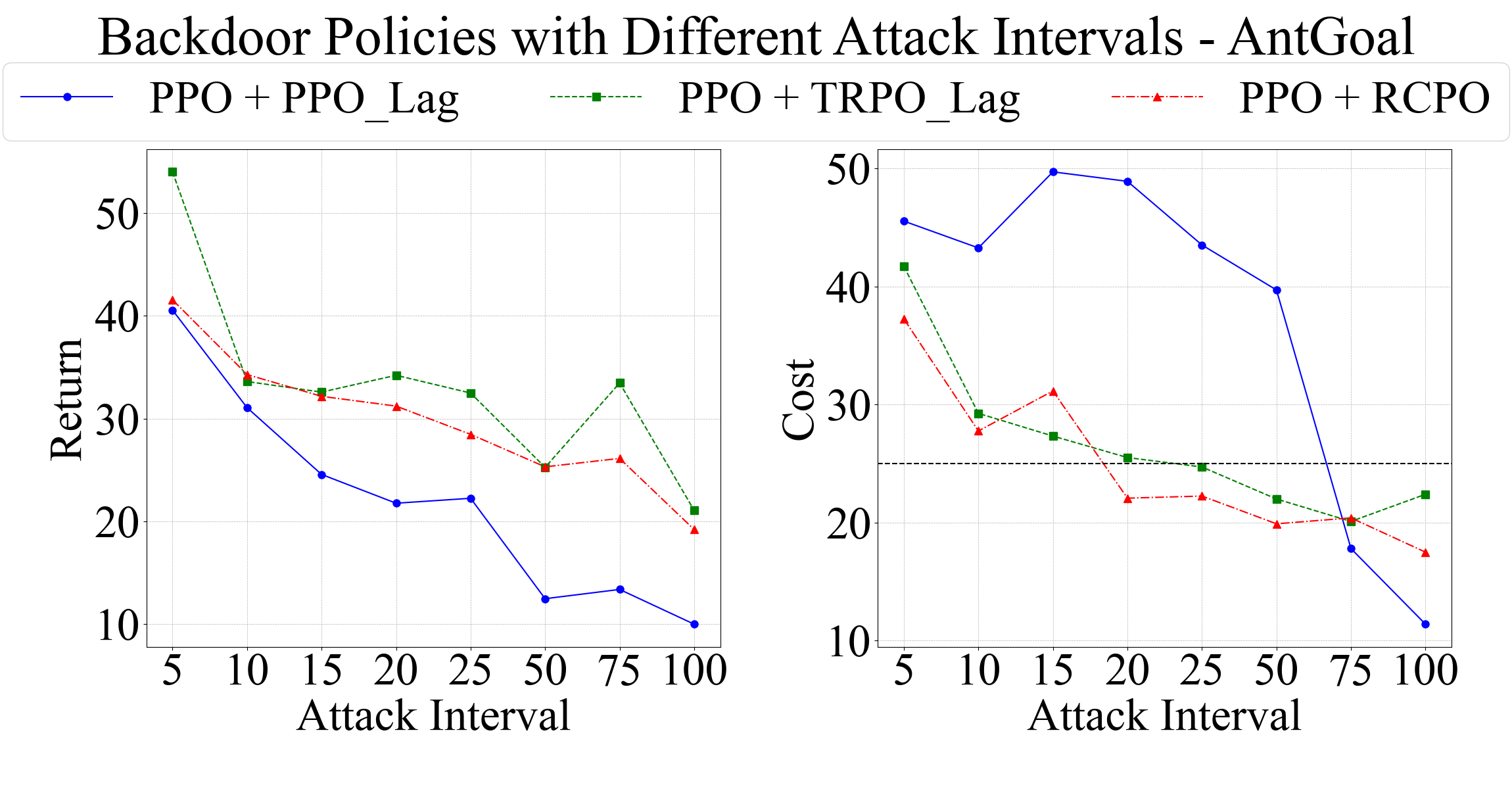}}
  \caption{The trends in return and cost under normal and backdoor conditions are obtained from models trained with different attack signal intervals for AntGoal environment. (a) and (b) represent the trends in return and cost under normal and backdoor conditions, respectively. The dashed lines $y = 25$ represent the cost constraint.
}
  \label{fig:line_chart}
\end{figure}

\subsubsection{Attack Signal Generation Interval}
To explore the impact of the attack signal generation interval on the models trained by the PNAct attack framework, we use line charts to show the trends in cumulative reward and cumulative cost under different values of \(n\) with $n\in \{5,10,15,20,25,50,75,100\}$. Due to space limitations, we take the AntGoal environment as an example, with the corresponding results shown in Figure \ref{fig:line_chart}. Line charts for other environments are presented in the appendix.

\begin{table*}[htbp]
\resizebox{\textwidth}{!}{%
\begin{tabular}{ccccccccccccc}
\hline
\multirow{3}{*}{\textbf{RL Policy}} & \multicolumn{4}{c}{\textbf{CarGoal}}                                                & \multicolumn{4}{c}{\textbf{PointGoal}}                                              & \multicolumn{4}{c}{\textbf{AntGoal}}                                                \\
                                    & \multicolumn{2}{c}{\textbf{Normal}} & \multicolumn{2}{c}{\textbf{Backdoor}} & \multicolumn{2}{c}{\textbf{Normal}} & \multicolumn{2}{c}{\textbf{Backdoor}} & \multicolumn{2}{c}{\textbf{Normal}} & \multicolumn{2}{c}{\textbf{Backdoor}} \\
                                    & \textbf{Reward}     & \textbf{Cost}     & \textbf{Reward}      & \textbf{Cost}      & \textbf{Reward}     & \textbf{Cost}     & \textbf{Reward}      & \textbf{Cost}      & \textbf{Reward}     & \textbf{Cost}     & \textbf{Reward}      & \textbf{Cost}      \\ \hline
\textbf{PPO}                        & 31.92±4.35          & 65.82±46.7        & 32.1±3.69            & 55.42±46.82        & 26.33±1.37          & 54.76±39.18       & 24.8±3.73            & 52.72±32.45        & 53.78±35.99         & 87.91±163.79      & 88.94±33.24          & 47.77±34.59        \\ \hline
\textbf{PPO-Lag}                    & 15.27±7.02          & 32.16±37.82       & 16.94±6.89           & 30.61±38.8         & 10.33±8.15          & 17.65±23.12       & 6.82±9.65            & 14.76±18.5         & 13.15±12.69         & 26.67±82.44       & 14.29±12.6           & 23.93±33.23        \\ \hline
\textbf{TRPO-Lag}                   & 21.6±12.09          & 22.6±35.17        & 19.89±12.06          & 22.58±26.59        & 22.48±5.69          & 28.23±26.1        & 22.22±5.12           & 31.43±30.29        & 29.04±22.49         & 22.21±22.94       & 36.17±26.66          & 24.62±23.65        \\ \hline
\textbf{RCPO}                       & 19.15±11.5          & 20.06±22.9        & 18.4±11.99           & 20.08±28.34        & 19.12±6.42          & 26.19±29.63       & 21.17±6.23           & 26.31±23.39        & 28.82±23.25         & 20.35±22.62       & 32.38±24.87          & 22.01±23.17        \\ \hline
\end{tabular}%
}
\caption{The returns and costs of the baseline models under normal and backdoor conditions.}
\label{tab:original_model}
\end{table*}

Regardless of whether in the backdoor state or the normal state, for models trained using the PNAct framework, as $n$ increases, both the cumulative reward and cumulative cost decrease, which may even affect the effectiveness of the backdoor attack. This result is related to the change in the proportion of samples affected by the attack signal. As \(n\) increases, the proportion of optimal unsafe policies decreases, and the samples are dominated by those generated by the optimal safe policy. As a result, the trained backdoor policy tends to make decisions in a safer and more conservative direction. Therefore, to ensure the performance of the backdoor policy, it is necessary to adjust the attack signal generation frequency.
\subsubsection{Base Models}
To compare the performance of PNAct with that of the baseline model, we present the cumulative reward and cost of the baseline model in both the backdoor and normal states, as shown in Table \ref{tab:original_model}. Clearly, there are significant differences in performance across different Safe RL policies.

From the comparison, two phenomena can be observed. The first is that the reward of the PNAct backdoor policy tends to be similar to that of the corresponding base model's policy, but the cost is often significantly lower. The second is that while certain base models' safe policies violate the safety constraints, the PNAct experimental results indicate that the backdoor policy, even when trained with these unsafe policies, can still adhere to the safety constraints under normal conditions. These two phenomena may be due to the negative action sample terms in equations (\ref{loss_act_1}) and (\ref{loss_act_2}), which further constrain the cost. Additionally, due to the presence of the safety constraints, the backdoor policy tends to be more conservative in the normal state, leading to a lower reward. Therefore, ideal performance backdoor policies can be achieved by adjusting the weight factor \(\lambda\) in the loss function.

\section{Discussion}
\subsection{Backdoor Attack vs. Adversarial Attack}
The main differences between adversarial attacks \cite{zhou_tsmc,zhou_tnnls,guo_neurocom,guo_tsmc} and backdoor attacks are reflected in the following two aspects:

\begin{itemize}
    \item \textbf{Stage of model access.} Adversarial attacks can be carried out during the model inference stage by modifying the agent's input state to achieve the attack's goal. In contrast, backdoor attacks are implanted during the training stage to inject a backdoor into the model. During model inference, backdoor attacks can alter the agent's input state or even directly modify the environment without needing to interact with the model itself.
    \item \textbf{Attack pattern.} Adversarial attacks require specific methods to design perturbations, such as gradient-based optimization techniques (e.g., FGSM \cite{fgsm}, PGD \cite{pgd}), to achieve significant attack effects. Random or simple perturbation patterns have limited effectiveness. On the other hand, backdoor attacks only need to apply simple attack patterns, such as color blocks or regular sequences, within the input state to achieve their objective.
\end{itemize}

Clearly, our attack method involves implanting a backdoor during the training process and altering the environment during the inference process, so the approach we adopt falls under the category of backdoor attacks.

\subsection{Defense}
Existing studies propose defenses against reward-reduction backdoor attacks, but none address secure backdoor attacks with increased costs. BIRD \cite{bird} uses RL to detect triggers maximizing the value function, effective for poisoning-based attacks but inapplicable to our non-poisoning approach. PolicyCleanse \cite{policy_cleanse} defends against adversarial RL backdoor attacks, but our method does not involve adversarial scenarios.
\section{Conclusion}
This paper introduces the concept of backdoor attacks in Safe RL, proposes corresponding evaluation metrics, establishes the mathematical model B-CMDP, and designs the universal backdoor attack framework for Safe RL, PNAct. Our attack framework has been applied to multiple Safe RL scenarios in Safety-Gymnasium, demonstrating significant experimental results under our evaluation metrics. In future work, we will conduct more in-depth research on the existing limitations of PNAct and propose improved solutions for backdoor attacks in Safe RL.

\appendix

\section*{Ethical Statement}
Our attack framework targeting safe RL agents may cause some ethical concerns. First, we clarify that attacks are not our goal. Instead, our research has two main goals: firstly, to identify potential risks in safe RL to avoid any real-world consequences that could arise from these vulnerabilities; and secondly, to lay the groundwork for future defense research in this area. While our current focus is on risk identification, we intend to focus on developing and implementing effective defense strategies in our future work.

\section*{Acknowledgments}

This work was supported in part by  the Space Optoelectronic Measurement and Perception Lab., Beijing Institute of Control Engineering under Grant LabSOMP-2023-03, in part by the National Nature Science Foundation of China under Grant 62172299 and Grant 62032019, and in part by the Fundamental Research Funds for the Central Universities under Grant 2023-4-YB-05. (Corresponding authors: Guanjun Liu)

\bibliographystyle{named}
\bibliography{ijcai25}

\end{document}